\documentclass[11pt]{article}
\usepackage[utf8]{inputenc}
\usepackage[T1]{fontenc}
\usepackage[margin=1in]{geometry}
\usepackage{booktabs,array,amsmath,graphicx,xcolor,hyperref,tikz}
\usetikzlibrary{arrows.meta,positioning}
\hypersetup{colorlinks=true,linkcolor=blue!50!black,citecolor=blue!50!black,urlcolor=blue!50!black}
\newcommand{\sql}[1]{\texttt{#1}}

\title{SqlToViz: A Contract-Pinned, Portable Pipeline for Recommending Charts\\ from SQL Queries and Result Sets\\[2pt]\large Preliminary evaluation against LLM-annotator agreement and nvBench; no human validation}
\author{Juan Manuel Castillo Pinto\\ Universidad de La Salle, Bogot\'a, Colombia\\ \texttt{ORCID 0009-0002-5751-0709}\thanks{Repository: \url{https://github.com/jmmana/sqltoviz}.}}
\date{2026-11-28}

\begin{document}
\maketitle

\begin{abstract}
Chart recommenders usually start from a table, a corpus of charts, or a natural-language request. Many applications, however, hold only a SQL query and its result set. We describe SqlToViz, a pipeline that recommends one of ten base chart classes (expanded deterministically to twenty visualizations plus a table fallback) from the query text and the executed result. The pipeline is built around a frozen, versioned feature contract (16 lexical SQL features, 16 table-level data features, 28 per-column features), a small ONNX classifier, hard semantic constraints that fail closed to a table, and cross-runtime parity fixtures so that Python, TypeScript and .NET implementations compute identical features. We report what has been measured: a leakage-audited synthetic benchmark; a held-out evaluation of 100 real queries against the \emph{agreement of LLM annotators} (three models and one AI-filled workbook of unknown model, not independent of one another); an external evaluation on nvBench, whose labels are human-authored; a feasibility study of role supervision; and parity, latency and numerical-faithfulness measurements. \textbf{There is no human validation of the recommendations}: the reference of the 100-item evaluation is LLM agreement, which is not human judgement. On the external human-authored nvBench labels the experimental models are \textbf{not validated and perform poorly on accuracy} (0.249 for the shipped constrained model against 0.763 for always predicting bar); only macro-F1 is above the baseline. We make no claim of novelty, of superiority over prior systems, or of agreement with human preferences, and the released models remain experimental candidates.
\end{abstract}

\section{Introduction}
Visualization recommendation has been studied through rule systems and design constraints \cite{mackinlay2007showme,wongsuphasawat2016voyager,moritz2018draco,yang2023draco2}, models learned from corpora of datasets and charts \cite{hu2019vizml,luo2018deepeye,zhou2020table2charts,dibia2019data2vis}, and natural-language-to-visualization translation \cite{luo2021nvbench,luo2022ncnet,dibia2023lida}; a survey is given by Zhu et al.\ \cite{zhu2020survey}. Our setting differs in its input: the system sees the SQL text and the rows the query returned, and it must be embeddable in application back-ends written in several languages. This motivates three design goals: (i) a feature contract that is frozen and hashed, so that independent implementations can be tested against each other; (ii) a hierarchical decision in which a learned model ranks \emph{base} candidates while deterministic rules specialize and validate them; and (iii) an evaluation protocol that keeps synthetic, LLM-generated and human labels strictly apart.

This paper is a progress report. We state hypotheses and a protocol (Section~\ref{sec:protocol}), present the measurements that exist, and mark clearly what is missing, in particular any validation by human annotators. We do not claim novelty; the literature review rests on full texts of eleven works and on abstracts of a few others.

\section{Related work}
We read the full texts of eleven closely related works (the twelfth, Show Me \cite{mackinlay2007showme}, was not accessible and is cited at title level only); notes with the exact figures are in the repository. Rule-based systems encode design knowledge: Voyager supports faceted browsing of automatically recommended views and reports a user study without a ranking-accuracy evaluation \cite{wongsuphasawat2016voyager}; Draco formalizes design knowledge as hard and soft constraints and learns soft-constraint weights from pairwise human judgments \cite{moritz2018draco}, and Draco~2 is a successor platform with a more flexible specification format, tests and APIs \cite{yang2023draco2}; Vega-Lite is the grammar many of these systems target \cite{satyanarayan2017vegalite}. Learned recommenders start from a table: VizML predicts per-choice design decisions (chart type among bar, line and scatter in its two- and three-class tasks, axis assignment) from engineered features of 119{,}815 deduplicated Plotly datasets and reports accuracy on class-balanced sets \cite{hu2019vizml}; DeepEye ranks four chart types using a learned good/bad classifier trained on 42 labelled tables \cite{luo2018deepeye}; Table2Charts generates chart-template sequences for whole fields without aggregation and reports recall of user-made charts on a 266{,}252-chart spreadsheet corpus \cite{zhou2020table2charts}; Data2Vis translates JSON rows into Vega-Lite specifications and reports log perplexity and qualitative failures \cite{dibia2019data2vis}. Natural-language approaches start from a question: nvBench synthesizes (NL, visualization) pairs from Spider's (NL, SQL) pairs, so its gold charts are valid synthesized candidates rather than a unique best chart \cite{luo2021nvbench}; ncNet translates language to Vega-Zero with a transformer \cite{luo2022ncnet}; LIDA is an LLM pipeline evaluated by visualization error rate \cite{dibia2023lida}. Recent work on benchmarks questions whether text-to-vis benchmarks reflect real use \cite{nguyen2024benchmarks}; see also VisEval \cite{chen2024viseval}, Prompt4Vis \cite{li2024prompt4vis} and the task-oriented TaskVis \cite{shen2022taskbased} (read at abstract level only). A survey is given by Zhu et al.\ \cite{zhu2020survey}. Reported metrics differ across these works (balanced accuracy, recall of user-made charts, tree-match accuracy on language input, NDCG, user-study counts), so we make no numeric comparison with them. To our reading of these works none takes the SQL text together with the executed result set, nor combines a learned ranker with hard eligibility constraints that fail closed; this is an observation, not a novelty claim. Spider \cite{yu2018spider} and BIRD \cite{li2023bird} are text-to-SQL resources; we use only their queries and databases as a source of realistic result sets. We could not compare against VizML directly because its data are inaccessible and carry no declared license (Section~\ref{sec:data}).

\section{Problem and taxonomy}
The model chooses among ten base candidates (\sql{kpi}, \sql{bar}, \sql{horizontal\_bar}, \sql{stacked\_bar}, \sql{line}, \sql{area}, \sql{donut}, \sql{scatter}, \sql{heatmap}, \sql{table}). A deterministic specializer expands them into twenty visualization types plus the table fallback (for example \sql{gauge} from \sql{kpi}, \sql{waterfall} and \sql{histogram} from \sql{bar}), and hard semantic constraints reject recommendations whose preconditions cannot be established (additivity, hierarchy, units, target). Unknown semantics fail closed; SQL syntax alone is never taken as proof of additivity. Model scores are probabilities over base classes, not calibrated confidences of the final chart.

\section{Method}
\subsection{Feature contract}
SqlVizSpec-v1.0.0 fixes 60 ordered features: 16 lexical SQL features (token count, joins, grouping, ordering, limits, aggregates, window and temporal functions, \dots), 16 table-level data features (row and column counts, type counts, non-null density, nominal cardinalities and entropy, a crossed-structure indicator), and 28 per-column features (type indicators, cardinality, string length, moment statistics, monotonicity, temporal-step irregularity, range, IQR and MAD normalizations, and context copies of table features). The contract also fixes the lexer, canonical types, null handling, binary64 intermediates with float32 outputs, and a parity tolerance of $10^{-6}$. The contract file is identified by its SHA256 (\texttt{d14d140b\dots5864}); nine ambiguous readings were documented and confirmed as normative by the contract owner, and the interpretation document is pinned in every report. The 32-dimensional table vector (SQL then global features) is the chart model's only input; the 28-dimensional column vectors feed the role model.

\subsection{Pipeline}
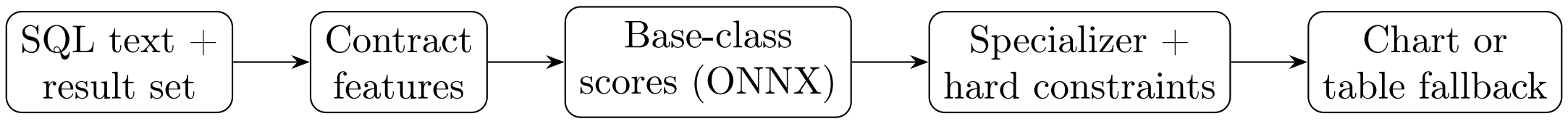
\begin{figure}[h]\centering
\begin{tikzpicture}[node distance=7mm, every node/.style={draw, rounded corners, align=center, font=\small, minimum height=8mm}, >=Stealth]
\node (a) {SQL text +\\result set};
\node (b) [right=of a] {Contract\\features};
\node (c) [right=of b] {Base-class\\scores (ONNX)};
\node (d) [right=of c] {Specializer +\\hard constraints};
\node (e) [right=of d] {Chart or\\table fallback};
\draw[->] (a)--(b); \draw[->] (b)--(c); \draw[->] (c)--(d); \draw[->] (d)--(e);
\end{tikzpicture}
\caption{Pipeline. The learned component is a classifier over the 32 table features; specialization and validation are deterministic and explain themselves with reason codes.}
\end{figure}
The chart model is a random forest (300 trees in experiments, 100 in the exported candidate, balanced class weights, fixed defaults, no tuning), exported to ONNX with inputs and outputs named by the contract. Training is single-threaded and the graph name is fixed, which makes the exported file byte-identical across independent re-trainings (verified).

\section{Data and labels}\label{sec:data}
\paragraph{Sources and licenses.} Spider and BIRD (CC BY-SA 4.0) supply SQL and databases; TPC-H/TPC-DS samples are generated locally with DuckDB (the TPC license restricts how results are disclosed, so none are presented as TPC results); VizML was excluded because its data were not downloadable and carry no license; nvBench declares MIT in its README without a license file and is not redistributed. Material of different licenses is kept in separate namespaces.
\paragraph{Execution and splits.} Gold queries run in a read-only sandbox with timeouts and row caps. Of 9{,}253 executable Spider/BIRD queries, 5{,}781 remain after within-database near-duplicate removal. Splits are grouped by database; we found that \sql{formula\_1} occurs in both Spider train and BIRD dev, so groups use the canonical database name.
\paragraph{Synthetic labels.} Twenty thousand synthetic items (2{,}000 per class) are generated from class-specific data regimes with class-agnostic schemas and SQL templates. Labels follow our own generation convention and therefore measure agreement with that convention, not human preference. A leakage audit (grouped cross-validation by schema) shows that SQL-only models reach accuracy 0.570 (SQL features) and 0.589 (token bag), and names-only models 0.509, against a structural ceiling of 0.60 (perfect knowledge of the result shape with uniform guessing inside each shape); within every shape their accuracy is at chance.
\paragraph{LLM pilot.} Three hundred Spider queries were labelled blind by two LLM annotators from different model families (percent agreement 92.7\%, Cohen's $\kappa=0.844$). One hundred stratified items were reserved for human validation and excluded from every training set and from the pilot experiments. The human validation was cancelled (Section~\ref{sec:protocol}); these items are instead evaluated against LLM agreement (Section~\ref{sec:llmref}). LLM agreement is never reported as human consensus.

\section{Hypotheses and protocol}\label{sec:protocol}
Eight research questions with falsifiable hypotheses (feature value of SQL versus data, effect of constraints, ablation, synthetic-to-real transfer, parity, cost, role learnability, anomaly faithfulness) and the confirmatory protocol are recorded in the repository (\texttt{research/paper/RESEARCH\_QUESTIONS.md}, \texttt{PROTOCOL.md}). The protocol fixes metrics (macro-F1 over classes with at least five reference items, cluster bootstrap over databases), the reserved items as the held-out set, and a list of analyses already seen and therefore exploratory. \textbf{Deviation D1 (project owner, 2026-11-28):} the protocol planned a confirmatory evaluation against at least two independent human annotators; the owner decided that no human validation will be collected. The reference is therefore the agreement of LLM annotators, declared as such; every output carries the provenance \texttt{llm\_reference} and \texttt{confirmatory: false}, hypotheses H1 and H4 are tested only against LLM agreement, and nothing here is called confirmatory.

\section{Experiments}
Every number below states what it measures. The only human-authored labels are those of nvBench (Section~\ref{sec:nvbench}); no human judged the output of our system.
\subsection{E1: synthetic training, held-out synthetic schemas}
Table~\ref{tab:e1} (three split seeds) measures agreement with the synthetic convention on unseen schemas. Rules were written by the same authors as the generator, so their lead partly reflects a shared convention. The table-level features cannot separate \sql{bar} from \sql{horizontal\_bar} (label length is a column feature) or \sql{line} from \sql{area} (sign and baseline are column features): in one split the hybrid model's F1 is $(0.82, 0.73)$ for (\sql{bar}, \sql{horizontal\_bar}) and $(0.46, 0.53)$ for (\sql{line}, \sql{area}), while the rules, which read column features, reach $(1.0, 1.0)$ and $(0.58, 0.79)$. This is a limitation of the contract's chart-model input.
\begin{table}[h]\centering\small
\begin{tabular}{lccc}\toprule
System & macro-F1 (mean $\pm$ sd) & accuracy & semantic violation rate\\\midrule
Rules only & $0.913\pm0.001$ & 0.919 & 0\\
SQL features only & $0.570\pm0.004$ & 0.573 & 0.085\\
Data features only & $0.855\pm0.004$ & 0.855 & 0.021\\
Hybrid (32 features) & $0.855\pm0.001$ & 0.855 & 0.022\\
Hybrid + hard constraints & $0.876\pm0.002$ & 0.877 & 0\\\bottomrule
\end{tabular}
\caption{E1, 20{,}000 synthetic items, grouped split by schema (14{,}000 train / about 3{,}000 test).}\label{tab:e1}
\end{table}
\paragraph{Ablation.} Leaving out one feature group at a time changes the hybrid macro-F1 (0.854) by at most 0.019 (data-size features $-0.019$, data-cardinality $-0.002$, SQL-size $+0.003$, other groups 0.000), i.e.\ the features are highly redundant on this data (exploratory).

\subsection{E2: transfer to real queries judged by LLMs}
Trained on all synthetic data and evaluated on the 131 LLM-consensus pilot items that survive deduplication (95 \sql{table}, 27 \sql{kpi}, 7 \sql{bar}, 2 \sql{horizontal\_bar}), Table~\ref{tab:e2} shows that unconstrained models over-recommend charts for real detail-style queries and that constraints remove these violations; the constrained hybrid is only slightly above the always-\sql{table} baseline (0.725). Only four classes occur in the pilot, so this experiment cannot assess the other six.
\begin{table}[h]\centering\small
\begin{tabular}{lccccc}\toprule
System & accuracy & macro-F1 (4 classes) & violations & table fallback\\\midrule
Rules only & 0.885 & 0.639 & 0 & 0.679\\
SQL only & 0.573 & 0.494 & 0.412 & 0.366\\
Data only & 0.229 & 0.221 & 0.656 & 0.015\\
Hybrid & 0.588 & 0.492 & 0.321 & 0.374\\
Hybrid + constraints & 0.908 & 0.644 & 0 & 0.695\\\bottomrule
\end{tabular}
\caption{E2 against LLM-consensus labels (not human), $n=131$, majority baseline 0.725.}\label{tab:e2}
\end{table}

\subsection{Reserved 100 queries against the LLM reference}\label{sec:llmref}
\textbf{Design.} No human labelled these items. The reference is the plurality label over four annotators: two blind pilot runs (Claude and a GPT model), a Claude single pass, and an AI-generated returned workbook of unknown model. The four are \emph{not independent} (two are the same model family, one is unidentified) and none is human; their agreement shows that the labels are reproducible by LLMs, not that they are right. Pairwise percent agreement on the primary label is 92--97\% (Cohen's $\kappa$ 0.83--0.93); 91 items are unanimous, 6 are decided 3--1, 1 is decided 2--1--1 and 2 ties are listed and excluded, leaving $n=98$. Systems are trained on the synthetic data only. Uncertainty is a paired cluster bootstrap over databases (2000 resamples, 82 clusters). All outputs are exploratory (\texttt{llm\_reference}, \texttt{confirmatory: false}).

\textbf{Class balance.} The reference contains \sql{table} 70 (71\%), \sql{kpi} 20, \sql{bar} 6 and \sql{horizontal\_bar} 2 items; the other six classes do not occur. Always answering \sql{table} reaches accuracy 0.714, so accuracy mostly measures how often a system answers \sql{table}. Macro-F1 is computed over the three classes with at least five items (\sql{table}, \sql{kpi}, \sql{bar}); \sql{horizontal\_bar} is excluded.
\begin{table}[h]\centering\small
\begin{tabular}{lcccc}\toprule
System & accuracy & macro-F1 [95\% CI] & acceptable-set hit & violations\\\midrule
Rules only & 0.888 & 0.636 [0.60, 0.67] & 0.969 & 0\\
SQL features only & 0.602 & 0.590 [0.47, 0.73] & 0.643 & 0.357\\
Data features only & 0.235 & 0.221 [0.16, 0.27] & 0.449 & 0.643\\
Hybrid (32 features) & 0.551 & 0.422 [0.35, 0.49] & 0.765 & 0.347\\
Hybrid + hard constraints & 0.888 & 0.634 [0.59, 0.66] & 0.959 & 0\\
Always \sql{table} & 0.714 & 0.277 [0.26, 0.30] & --- & ---\\\bottomrule
\end{tabular}
\caption{Reserved items ($n=98$) against the LLM-reference plurality (not human). Acceptable-set hit uses the union of the annotators' acceptable charts and is lenient.}\label{tab:llmref}
\end{table}
Paired differences whose 95\% interval excludes zero: rules and the constrained hybrid are above data-only, hybrid and always-\sql{table}; the constraints raise the hybrid by 0.215 [0.15, 0.28]. Rules and the constrained hybrid are indistinguishable (0.003 [0.00, 0.01]), as are rules and SQL-only; all 15 pairwise differences are in the report. The constrained systems behave as ``\sql{table} unless the result has one row'': the constrained hybrid recommends \sql{donut} for all six \sql{bar} items (and so obtains \sql{bar} F1 of 0), and its high acceptable-set hit only reflects that annotators often accept a donut. Scored against each annotator separately, the constrained-hybrid macro-F1 over supported classes is 0.69, 0.55, 0.68 and 0.63 (pilot Claude, pilot Codex, Claude single pass, AI workbook), so the result depends on which LLM is taken as reference. With 6 and 2 items in the rare classes these figures say little about any class other than \sql{table} and \sql{kpi}.

\textbf{LLM-as-recommender baseline.} The harness (versioned prompt, strict parser, per-item response log, offline replay) is implemented and tested, but \textbf{it was not run}: no provider API key was available. It needs an Anthropic or OpenAI key, network access and a model identifier. We therefore report no LLM-recommender numbers; a Claude or GPT model scored against this reference would in any case be circular, since such models are among the annotators.

\subsection{External evaluation on nvBench}\label{sec:nvbench}
nvBench provides human-authored (natural-language question, chart) pairs over Spider databases. We run the released experimental models on 7{,}097 items, executing each item's SQL verbatim in the sandbox; items whose SQL fails (145) or is rejected by the contract (5) are dropped and counted. Pie maps to \sql{donut}, Grouping Line to \sql{line} and Grouping Scatter to \sql{scatter}; nvBench covers 5 of our 10 classes and 76\% of its items are \sql{bar}. The label is the chart \emph{requested in the question}, not necessarily the best chart for the data, and the engine never sees the question. The data are downloaded locally and are neither redistributed nor committed (the nvBench license is ambiguous).
\begin{table}[h]\centering\small
\begin{tabular}{lcc}\toprule
System & accuracy [95\% CI] & macro-F1, 5 classes [95\% CI]\\\midrule
Always \sql{bar} & 0.763 [0.730, 0.793] & 0.173 [0.169, 0.177]\\
Model argmax & 0.286 [0.223, 0.371] & 0.342 [0.286, 0.389]\\
Model + hard constraints (shipped) & 0.249 [0.180, 0.345] & 0.220 [0.168, 0.286]\\
Rules only & 0.275 [0.209, 0.358] & 0.159 [0.137, 0.186]\\\bottomrule
\end{tabular}
\caption{nvBench, SQL verbatim, $n=7097$, cluster bootstrap over 152 databases (exploratory; external human-authored labels).}\label{tab:nvbench}
\end{table}
\textbf{The models are not validated by this evaluation, and their accuracy is far below the majority baseline.} The shipped model's accuracy (0.249) is far below always-\sql{bar} (0.763); the difference is $-0.512$ [$-0.600$, $-0.410$]. Its macro-F1 is above the rules baseline by 0.060 [0.020, 0.111] and the raw argmax is above it by 0.180 [0.122, 0.235]; the shipped model's macro-F1 is \emph{not} distinguishable from always-\sql{bar} (difference 0.047 [$-0.006$, 0.114]). The models over-predict \sql{donut} and \sql{table}, almost never predict \sql{line} (F1 0.03 and 0.04), and the hard constraints remove \sql{stacked\_bar} entirely (F1 0.00) because nvBench's SELECT lacks the series column that stacked-bar eligibility needs; with nvBench's own chart data the constrained macro-F1 is 0.409 [0.352, 0.480]. Top-2 accuracy (0.38) is also below always-\sql{bar}. Classes \sql{kpi}, \sql{horizontal\_bar}, \sql{area}, \sql{heatmap} and \sql{table} are not assessed. Because the models were trained on synthetic labels only, this result argues for training and evaluation on real, human-labelled data, not for any quality claim; we do not claim an improvement over a majority baseline in accuracy.

\subsection{Role supervision feasibility}
No corpus used above carries role labels. From 817 Vega-Lite example specifications \cite{satyanarayan2017vegalite} (BSD-3-Clause, pinned commit), 136 have inline data and simple encodings, yielding 322 labelled columns. With grouped five-fold cross-validation by specification, a random forest on the 28 column features reaches macro-F1 0.50 (95\% cluster-bootstrap CI 0.44--0.56), versus 0.32 for a positional heuristic and 0.12 for the majority class. The set is small and imbalanced (10 examples for the size/secondary role) and not SQL-aware; this is exploratory feasibility evidence only: it measures agreement with the Vega-Lite authors' channel choice on held-out specifications, not the quality of any recommendation.

\subsection{Portability, parity and cost}
The candidate chart model (5.77 MB) matches scikit-learn on 2{,}942 held-out items with maximum absolute difference $2.6\times10^{-7}$ (tolerance $10^{-6}$) and identical argmax, and its export is byte-identical across independent re-trainings. The 32 table and 28 column features were implemented independently in Python, TypeScript and C\#. All three reproduce 308 golden fixtures (the original 288 plus 20 portability cases: astral symbols, non-ASCII digits and whitespace, integers written as floats, near-constant columns) within $10^{-6}$ with identical rejections, and 3{,}000 differential fuzz cases generated from the Python implementation (random schemas, missing values, quoted identifiers, window and CTE forms, malformed SQL; 2{,}225 accepted and 775 rejected, identical in every runtime). With the ONNX models, Python, Node (native), browser WebAssembly and .NET 8/10 reproduce the golden chart scores within $10^{-6}$ and give identical base and constrained classes, eligibility and column roles. Making this possible required explicit numerical rules, e.g.\ left-to-right summation (Python $\geq 3.12$ compensates sums), explicit products instead of powers, an exact constant-column test (otherwise a column of $0.1$ repeated gives kurtosis $-3$ in some runtimes and $0$ in others), and treating integral numbers as integers because JSON does not distinguish $1$ from $1.0$; these were reported to the contract owner as clarifications and do not change the contract bytes.

Table~\ref{tab:cost} gives end-to-end latency (feature extraction, ONNX inference and constraints) on one Apple M4 Pro with single-threaded ONNX Runtime; feature extraction dominates from about 1{,}000 rows, ONNX inference stays at 0.01--0.11 ms, constraints take at most 0.02 ms, and exact median/quantiles make extraction $O(NM\log N)$ (measured 9.5--10.7$\times$ per decade of rows). Cold engine loading takes 63--357 ms (Python 63, .NET 82, Node 113, Chromium WASM 357). These figures describe this hardware and these implementations only.
\begin{table}[h]\centering\small
\begin{tabular}{rcccc}\toprule
Rows & Python & Node & .NET & Chromium WASM\\\midrule
10 & 0.232 & 0.079 & 0.066 & 0.038\\
100 & 0.623 & 0.165 & 0.302 & 0.101\\
1{,}000 & 4.659 & 1.051 & 1.142 & 0.710\\
10{,}000 & 45.391 & 9.636 & 12.109 & 7.500\\\bottomrule
\end{tabular}
\caption{Median end-to-end latency in milliseconds (400 runs; 50 for 10{,}000 rows; the browser column times batches because of the 0.1\,ms timer resolution).}\label{tab:cost}
\end{table}

\subsection{Separating bar from horizontal bar and line from area}
The 32 table-level features cannot separate these pairs (Section E1), but the contract's column features can. A deterministic resolver applied after the model and the constraints (horizontal bar if the longest nominal label is long; area if all measures are non-negative and the series spans at least half of its mean) with thresholds chosen on a validation split raises held-out macro-F1 on the synthetic benchmark from 0.872 to 0.985 without changing the contract, while adding three column aggregates to the model input reaches 0.992. Because the generator defines these classes by the same cues, this shows that the information is available, not that people prefer these choices; the pilot contains no area and almost no horizontal bars to test it.

\subsection{Statistical detectors}
On synthetic Gaussian series with injected outliers, the MAD detector recovers all outliers with 10\% contamination and $6\sigma$ outliers at $n=100$ while a classical $z$-score detector recovers 5\%; with 2\% contamination at $4\sigma$ the $z$-score detector recalls more (0.99 versus 0.82). On clean series at $n=1000$, 38.8\% contain at least one flagged point, so results should be presented as candidates rather than alarms. Against exact rational and 120-digit decimal references on 1{,}500 random cases, the Python, TypeScript and .NET detectors agree in statuses, outlier sets and Pareto inclusion sets, with worst errors $5.3\times10^{-14}$ (MAD, relative), $\leq 6.7\times10^{-16}$ (Pearson, absolute), $2.2\times10^{-16}$ (temporal shifts) and $\leq 5.6\times10^{-16}$ (Pareto).

\section{Software release status}\label{sec:release}
Version 0.1.0-alpha.2 is published as an alpha pre-release in the three registries (verified against the registries on 2026-11-27): PyPI (\texttt{sqltoviz} 0.1.0a2), npm (\texttt{sqltoviz} 0.1.0-alpha.2, \texttt{alpha} tag) and NuGet (\texttt{SqlToViz} and \texttt{SqlToViz.Onnx} 0.1.0-alpha.2). The three packages carry the same frozen contract (SqlVizSpec-v1.0.0) and the same parity fixtures. The bundled ONNX chart model is an \emph{experimental candidate}: it is trained on synthetic labels, it has not been validated by human annotators (and, on nvBench's human-authored labels, it performs poorly in accuracy), and its recommendations should not be taken as evidence of quality for any particular application. The release is a software milestone, not a result of this paper.

\section{Limitations and threats to validity}
(1) \textbf{No human validation.} No human labelled the reserved items or judged any recommendation; the reference of Section~\ref{sec:llmref} is the agreement of four LLM annotators (three models and an AI workbook of unknown model) that are not independent, and LLM agreement is not human preference. (2) That reference is 71\% \sql{table}, contains four classes only (two with fewer than five items), and results depend on which annotator is the reference; macro-F1 on 98 items is unstable. (3) \textbf{On nvBench's human-authored labels the models are not validated and are poor in accuracy} (0.249 versus 0.763 for always \sql{bar}); nvBench covers five classes, labels the requested rather than the best chart, and is dominated by \sql{bar}. (4) Synthetic labels encode our conventions; rules and generator share authors. (5) Table-level features cannot distinguish two class pairs; a deterministic resolver fixes this only within the synthetic convention. (6) Role supervision is tiny and SQL-agnostic. (7) Real data come from Spider/BIRD gold queries, which are cleaner than production workloads. (8) Parity was verified on 308 golden and 3{,}000 fuzz cases generated by our own reference implementation, not on external implementations; cost was measured on one machine, in Chromium only, and .NET 8 only in continuous integration. (9) The protocol was frozen after several exploratory analyses and then deviated from (D1); Section~\ref{sec:protocol} lists both. (10) The released models are experimental candidates of an alpha release (Section~\ref{sec:release}); no recommendation-quality claim applies to them. (11) The LLM-as-recommender baseline was not run (no provider key); we report no numbers for it. (12) We make no novelty claim and no comparison with prior systems.

\section{Reproducibility}
Code, notebooks, manifests with SHA256 checksums, the contract and interpretation pins, byte-reproducible ONNX files, parity fixtures and all reports are in the repository (the scripts \texttt{run\_llm\_reference}, \texttt{run\_llm\_recommender}, \texttt{run\_roles} and \texttt{eval\_nvbench} under \texttt{research/kaggle/tools}); the Kaggle notebooks accept a pinned commit. Datasets derived from Spider/BIRD carry CC BY-SA 4.0 attribution and are kept separate from Apache-2.0 artifacts.

\section{Conclusion}
The engineering contract, tooling and measurement apparatus are in place. Against an LLM-agreement reference on 98 real queries the constrained systems largely answer \sql{table} or \sql{kpi}; against nvBench's human-authored labels the experimental models are not validated and their accuracy is far below a majority baseline. Whether the recommendations agree with human judgement remains unanswered; no human validation is planned, and the released models stay experimental candidates. Training and evaluation on real human-labelled data is the clear next step.

\bibliographystyle{plain}
\bibliography{references}
\end{document}